\documentclass[11pt]{article}
\usepackage{coling2020}
\usepackage{times}
\usepackage{hyperref}
\usepackage{latexsym}

\usepackage{graphicx} 
\usepackage{mathtools}
\usepackage{amssymb}
\usepackage{amsmath}

\colingfinalcopy %

\title{Leveraging Visual Question Answering to Improve Text-to-Image Synthesis}

\author{Stanislav Frolov \textsuperscript{1,2}, Shailza Jolly \textsuperscript{1,2}, J\"orn Hees \textsuperscript{2}, Andreas Dengel \textsuperscript{1,2} \\
\textsuperscript{1} Technical University of Kaiserslautern, Germany  \\
  \textsuperscript{2} German Research Center for Artificial Intelligence (DFKI), Germany \\
  
  {\tt firstname.lastname@dfki.de} \\
}

\date{}

\begin{document}
\maketitle
\begin{abstract}
Generating images from textual descriptions has recently attracted a lot of interest.
While current models can generate photo-realistic images of individual objects such as birds and human faces, synthesising images with multiple objects is still very difficult.
In this paper, we propose an effective way to combine Text-to-Image (T2I) synthesis with Visual Question Answering (VQA) to improve the image quality and image-text alignment of generated images by leveraging the VQA 2.0 dataset.
We create additional training samples by concatenating question and answer (QA) pairs and employ a standard VQA model to provide the T2I model with an auxiliary learning signal.
We encourage images generated from QA pairs to look realistic and additionally minimize an external VQA loss.
Our method lowers the FID from 27.84 to 25.38 and increases the R-prec. from 83.82\% to 84.79\% when compared to the baseline, which indicates that T2I synthesis can successfully be improved using a standard VQA model.
\end{abstract}

\begin{figure}[h]
\centering
\includegraphics[scale=0.8]{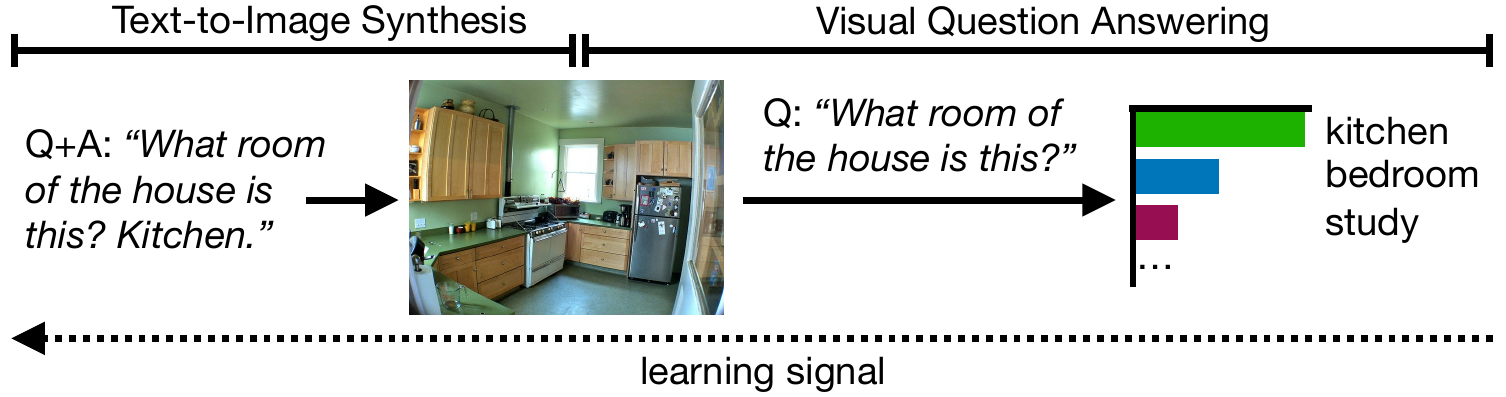}
\caption{
We use concatenated question answer (QA) pairs from VQA data as additional input samples for T2I training.
A standard VQA model can then be used to provide an additional loss to the T2I model.
}
\label{front-image}
\end{figure}

\NoHyper\blfootnote{
    
     \hspace{-0.65cm}  %
     This work is licensed under a Creative Commons 
     Attribution 4.0 International Licence.
     Licence details:
     \url{http://creativecommons.org/licenses/by/4.0/}.
}\endNoHyper

\section{Introduction}

\label{intro}
Text-to-image synthesis (T2I), the task to generate realistic images given textual descriptions, received a lot of attention in recent years \cite{Reed2016,StackGAN,StackGANv2,AttnGAN,DM-GAN,Obj-GAN}.
T2I synthesis can be seen as the inverse of image captioning. Given a caption, the model is trained to produce realistic images that correctly reflect the meaning of the input captions.
Many existing T2I methods use Generative Adversarial Networks (GANs) \cite{gan}.
GANs consist of two artificial neural networks that play a game in which a discriminator is trained to distinguish between real and generated images, while a generator is trained to produce images to fool the discriminator.
They have successfully been applied to many image synthesis applications such as image-to-image translation \cite{Isola_2017_CVPR,Zhu_2017_ICCV}, image super-resolution \cite{super-resolution}, and image in-painting \cite{image-inpainting}.

Similarly, Visual Question Answering (VQA) \cite{antol2015vqa} emerged as an important task to build systems that better understand the relationship between vision and language by learning to answer questions about an image.
It can be seen as image conditioned question answering, which encourages the model to both look at the image and understand the question to predict the correct answer.

A good T2I model should produce images that look realistic and correctly reflect the semantic meaning of the input description.
Considering the complexity of natural language descriptions and difficulty to produce photographic images, current models struggle to achieve these goals.
In this paper we propose a simple, yet effective way to combine T2I with VQA to improve both the image quality and image-text alignment of generated images of a T2I model by leveraging the questions and answers (QA) provided in the VQA 2.0 dataset \cite{antol2015vqa}.
Both captions and QA pairs can describe the overall image or very specific details.
In fact, many QA pairs can be rephrased as captions and vice versa which further motivates to leverage VQA for T2I.
Additionally, the VQA 2.0 dataset contains complementary images with different answers for the same question which
requires the T2I model to learn to pay close attention to the input in order to generate an image that correctly reflects the meaning of the input text.
To leverage the VQA 2.0 dataset for T2I, we concatenate QA pairs and use them as additional training samples in our T2I pipeline.
Images generated from QA pairs can subsequently be used as inputs to a VQA model which can provide an additional learning signal to the T2I generator.
See \autoref{front-image} for an overview of our approach.

\section{Related Work}
\label{related_work}

Initial T2I approaches \cite{Reed2016,Dash2017} adopted the conditional GAN (cGAN) \cite{cGAN} and AC-GAN \cite{ACGAN} ideas to replace the conditioning variable by a text embedding which allows to condition the generator on a textual description.
Analog to many current approaches, our approach is also based on AttnGAN \cite{AttnGAN}, which incorporates an attention mechanism on word features to allow the network to synthesize fine-grained details.

In terms of using VQA for T2I, to our knowledge the only other approach is VQA-GAN \cite{Niu2020ImageSF}. However, in contrast to their approach, our architecture is simpler, we do not use layout information, work on individual QA pairs, and produce higher resolution images (256x256 vs. 128x128).

\section{Approach}
\label{method}

\begin{figure}[t]
\centering
\includegraphics[scale=0.8]{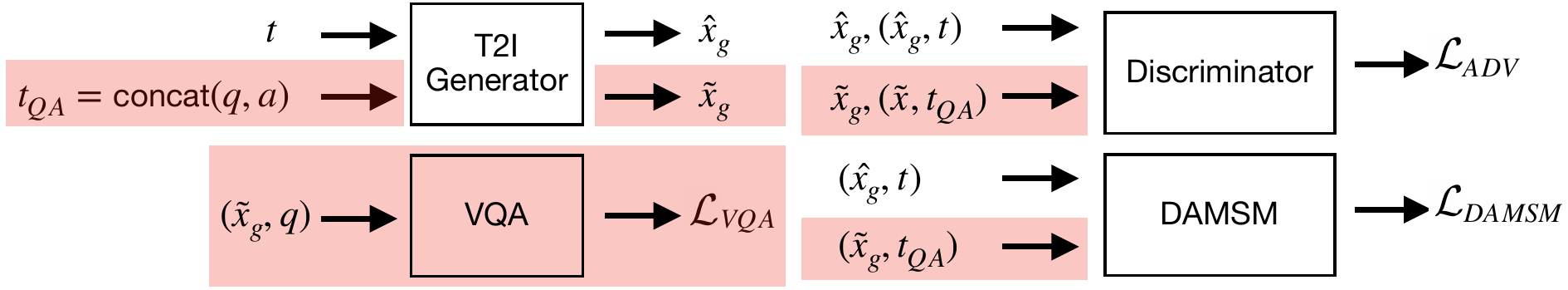}
\caption{
An overview of our architecture with our VQA extension highlighted in red. Given text inputs $t$ (captions) and $t_\textrm{QA}$ (concatenated QA pairs) we generate images $\hat{x}_g$ and $\tilde{x}_g$.
As in AttnGAN, we use the discriminator and DAMSM model to differentiate between real (omitted in the figure for brevity) and generated images as well as image-text pairs, leading to loss $\mathcal{L}_{\textrm{ADV}}$, and DAMSM loss $\mathcal{L}_{\textrm{DAMSM}}$ for improved image-text alignment.
Images generated from QA pairs are passed through a VQA model resulting in the additional loss $\mathcal{L}_{\textrm{VQA}}$.
}
\label{approach}
\end{figure}
We extend the AttnGAN \cite{AttnGAN} architecture to leverage question and answer (QA) pairs from the VQA 2.0 \cite{vqa_v2} dataset by appending a VQA model \cite{Kazemi2017ShowAA}.
In addition to generating images from image descriptions, our model is also trained to produce images given a QA pair and minimize an external VQA loss.
See Figure \ref{approach} for an overview.
After revisiting the individual components, we explain our extension in more detail.

\subsection{T2I: AttnGAN}
AttnGAN \cite{AttnGAN} consists of a multi-stage refinement pipeline that employs attention-driven generators for fine-grained T2I synthesis.
A pre-trained bidirectional LSTM (BiLSTM) \cite{Schuster1997BidirectionalRN} is used to extract global sentence as well as individual word features.
Discriminators jointly approximate the conditional and unconditional distributions simultaneously.
Additionally, they use an image-text matching loss at the word level called Deep Attentional Multimodal Similarity Model (DAMSM) is employed to guide the image generation process.
The attention-driven generator together with the DAMSM loss help the generator to be able to focus on individual words to synthesize fine-grained details and improve the semantic alignment between input description and final image.

\subsection{VQA Model}
We use a very basic and widely used VQA model, proposed in  \cite{Kazemi2017ShowAA} \footnote{\url{https://github.com/Cyanogenoid/pytorch-vqa}}.
Given an image $I$ and question $q$, the model uses image features extracted from a pre-trained ResNet \cite{He2016DeepRL}, an LSTM \cite{Hochreiter1997LongSM} based question embedding, and employs stacked attention \cite{yang2016stacked} to produce probabilities over a fixed set of answers.
The VQA loss, given in Equation~\ref{vqa_loss}, is simply an average over the negative log-likelihoods over all the correct answers $a_1,a_2,...,a_K$.

\begin{equation}
\label{vqa_loss}
  \mathcal{L}_{\textrm{VQA}}=\frac{1}{K}\sum_{k=1}^{K}-\textrm{log}P(a_k|I,q)
\end{equation}

\subsection{T2I + VQA}
We extend AttnGAN \cite{AttnGAN} by appending the VQA model \cite{Kazemi2017ShowAA}, and create additional training samples by concatenating question and answer (QA) pairs.
Given captions $t$ and QA pairs $t_{\textrm{QA}}$, we generate fake images $\hat{x}_g$, and $\tilde{x}_g$.
Real images are denoted as $x_r$.
Next, the VQA model takes the image generated from the QA pair and corresponding question to produce an answer.
Similar to \cite{Niu2020ImageSF}, we use the VQA loss to predict the correct answer to guide the image generator.
At the same time, the discriminators encourage the generated images to be realistic and matching to their corresponding text input.
The DAMSM loss \cite{AttnGAN} further helps to focus on the individual words to generate fine-grained details.
The final objective of our generator is defined in Equation~\ref{g_loss}, where $\mathcal{L}_\textrm{ADV}$ is the adversarial (conditional and unconditional) loss, and $\mathcal{L}_\textrm{DAMSM}$ is the DAMSM loss, both as described in  \cite{AttnGAN}.
The adversarial loss and DAMSM loss are applied to both captions and QA pairs and their correspondingly generated images.
The VQA loss $\mathcal{L}_\textrm{VQA}$ is applied only to images generated from QA pairs.

\begin{equation}
\label{g_loss}
\begin{split}
  \mathcal{L}_{G} =& \mathcal{L}_\textrm{ADV} + \mathcal{L}_\textrm{DAMSM}(\hat{x}_g,t) + \mathcal{L}_\textrm{DAMSM}(\tilde{x}_g,t_{\textrm{QA}}) + \mathcal{L}_{\textrm{VQA}}(\tilde{x}_g,q) \\
  \mathcal{L}_\textrm{ADV} =& \underbrace{-\mathbb{E}[\textrm{log}D(\hat{x}_g)]-\mathbb{E}[\textrm{log}D(\tilde{x}_g)]}_{\textrm{unconditional loss}} \underbrace{-\mathbb{E}[\textrm{log}D(\hat{x}_g,t)]-\mathbb{E}[\textrm{log}D(\tilde{x}_g,t_{\textrm{QA}})]}_{\textrm{conditional loss}}
\end{split}
\end{equation}

\noindent
The discriminators are trained to classify between real and fake images, as well as image-caption and image-QA pairs to simultaneously approximate the conditional and unconditional distributions by minimizing the modified loss defined in Equation~\ref{d_loss}.

\begin{equation}
\label{d_loss}
\begin{split}
  \mathcal{L}_{D}= & \underbrace{-\mathbb{E}[\textrm{log}D(x_r)]          -\mathbb{E}[\textrm{log}(1-D(\hat{x}_g))]         -\mathbb{E}[\textrm{log}(1-D(\tilde{x}_g))]}_{\textrm{unconditional loss}} \\
                     & \underbrace{-\mathbb{E}[\textrm{log}D(x_r, t)] -\mathbb{E}[\textrm{log}(1-D(\hat{x}_g,t))] -\mathbb{E}[\textrm{log}(1-D(\tilde{x}_g,t_{\textrm{QA}}))]}_{\textrm{conditional loss}}
\end{split}
\end{equation}

\section{Experiments}
\label{experiments}
We train three different variants of our extension.
The first two are naive extensions in which we do not change the discriminator loss of AttnGAN.
Instead, we simply append the VQA model, sample a QA pair during training and add the external VQA loss to the overall generator loss function.
We experiment with an end-to-end training approach where we start with a randomly initialized VQA model, and a pre-trained VQA model.
Next, we train a model in which we change the discriminator and generator loss functions.
In other words, given a QA pair, the model is not only trained to minimize the VQA loss for that particular QA pair, but also to produce realistic and matched images as judged by the discriminator and DAMSM loss.
For fair comparison we re-train and re-evaluate AttnGAN using the codebase provided by the authors\footnote{\url{https://github.com/taoxugit/AttnGAN}}.

\subsection{Datasets and Evaluation}
We use the commonly used COCO  \cite{coco} and VQA 2.0 \cite{vqa_v2} datasets to train our model.
COCO depicts complex scenes and multiple interacting objects and contains around 80k images for training and 40k images for testing.
Each image has five captions.
VQA 2.0 is a large dataset of question answer pairs based on the COCO images with roughly 400k QAs for training and 200k for testing, hence extensively used by VQA researchers \cite{tan2019lxmert,kim2018bilinear,lu2019vilbert}.
It contains complementary images for the same question, such that the model learns to look closely at the image before answering instead of deploying language biases \cite{antol2015vqa}.

We evaluate the quality and diversity of generated images using the Inception Score (IS) \cite{is} and Fréchet Inception Distance (FID) \cite{fid}.
To compute the IS\footnote{\url{https://github.com/sbarratt/inception-score-pytorch}} and FID\footnote{\url{https://github.com/mseitzer/pytorch-fid}} we generate 30k images from 30k randomly sampled test captions.
R-prec. \cite{AttnGAN} is used to evaluate the semantic alignment between generated images and input captions.
Although R-prec. might be unreliable, as current models seem to achieve higher scores than real images \cite{SOA}, we include it for reference since it is still commonly used.
Similar to \cite{Niu2020ImageSF}, we evaluate the VQA accuracy of our models by generating images from test QA pairs and passing them to the pre-trained VQA model with the corresponding question and answer.

\subsection{Results}

\begin{table}[th]
\begin{center}
\begin{tabular}{l|cccc}
\hline \bf Method& \bf IS $\uparrow$ & \bf FID $\downarrow$ & \bf R-prec. $\uparrow$ & \bf VQA Acc. $\uparrow$ \\ \hline
Real Images & 34.88 & 6.09 & 68.58 & 60.00 \\
\hline
AttnGAN \cite{AttnGAN} &  \bf 26.66 & 27.84 & 83.82 & 43.00 \\
AttnGAN + end-to-end VQA & 25.22 & 30.68 & 82.68 & 42.85 \\
AttnGAN + pre-trained VQA & 26.02 & 28.72 & 84.25 & 42.83 \\
AttnGAN + pre-trained VQA + adapted loss & \underline{26.64} & \bf 25.38 & \bf 84.79 & \bf 43.75 \\
\hline
\end{tabular}
\end{center}
\caption{Results on the COCO test dataset. First row contains scores for real images (excluding VQA Acc.) as reported in~\protect\cite{SOA}. We re-train and re-evaluate the baseline AttnGAN (second row). Third and fourth row are our naive extensions in which we simply append a VQA model for an external VQA loss for images generated from QA pairs. In the last row, we change the discriminator and generator losses to also encourage images generated from QA pairs to look realistic and match the input (similar to standard AttnGAN losses for images generated from captions). We train each model for 120 epochs, select the checkpoint with the best IS and report corresponding FID, R-prec., and VQA accuracy.}
\label{results}
\end{table}

As can be seen in Table~\ref{results}, merely adding the external VQA loss to the generator impairs the performance, regardless of using a pre-trained VQA model or training it as part of the pipeline in an end-to-end way.
We hypothesize this is due to the images produced from QA pairs not being encouraged to look realistic during training and the generator struggling to minimize the external VQA loss for images generated from QA pairs, on the one hand, and standard AttnGAN losses for images generated from captions, on the other hand.

Therefore, in our third experiment (last row) we change the overall objective and encourage all generated images to be realistic and matched to the input description using the conditional and unconditional losses from the discriminator and DAMSM loss (as in standard AttnGAN).
While achieving almost identical IS, our model greatly improves the FID from 27.84 to 25.38, which indicates better image quality and diversity.
Additionally, the images produced by our extension are better aligned to the input descriptions as indicated by the improvement of R-prec. from 83.82 to 84.79, and VQA accuracy from 43.00 to 43.75.
Since the VQA 2.0 dataset contains complementary image and QA pairs, the slight variation in linguistic inputs might hence help the model to generate better images.
Our results show that it is possible to improve T2I synthesis by simply appending a standard pre-trained VQA model, leveraging the VQA 2.0 dataset as additional supervision and encouraging the model to also produce realistic images from QA pairs.
Although we show that already a simple VQA model can help to improve the final T2I performance, we hypothesize that using a state-of-the-art VQA model might further improve the results.

\section{Conclusion}
\label{conclusion}
In this paper we proposed a simple method to leverage VQA data for T2I via a combination of AttnGAN, a well-known T2I model, and a standard VQA model.
By concatenating quesion answer (QA) pairs from the VQA 2.0 dataset we created additional training samples.
Images generated from the QA pairs are passed to the VQA model which provides an additional learning signal to the generator.
Our results show a substantial improvement over the baseline in terms of FID, but requires additional supervision.
Possible future research directions could be to investigate whether our extension also boosts other T2I models, and the influence of more training data and the external VQA loss separately.

\section*{Acknowledgements}

This work was supported by the BMBF project DeFuseNN (Grant 01IW17002) and the TU Kaiserslautern PhD program.

\bibliographystyle{coling}
\bibliography{coling2020}

\end{document}